\documentclass{bmvc2k}
\usepackage{siunitx}
\DeclareMathAlphabet{\mathpzc}{OT1}{pzc}{m}{it}
\title{Y-Net: A deep Convolutional Neural Network \\ for Polyp Detection}

\addauthor{Ahmed Mohammed}{ mohammed.kedir@ntnu.no}{1}
\addauthor{Sule Yildirim}{sule.yildirim@ntnu.no}{2}
\addauthor{Ivar Farup}{ivar.farup@ntnu.no}{1}
\addauthor{Marius Pedersen}{marius.pedersen@ntnu.no}{1}
\addauthor{{\O}istein Hovde }{Oistein.Hovde@sykehuset-innlandet.no}{3}

\addinstitution{
Department of Computer Science,\\
 NTNU,\\
Gj{\o}vik, Norway
}
\addinstitution{
Department of Information Security and Communication Technology,\\
 NTNU,\\
Gj{\o}vik, Norway
}

\addinstitution{
	Department of Gastroenterology,\\
University of Oslo, \\
	Oslo, Norway
}

\runninghead{Y-Net}{A deep Convolutional Neural Network for Polyp Detection}

\begin{document}

\maketitle

\begin{abstract}
Colorectal polyps are important precursors to colon cancer, the third most common cause of cancer mortality for both men and women. It is a disease where early detection is of crucial importance. Colonoscopy is commonly used for early detection of cancer and precancerous pathology. It is a demanding procedure requiring significant amount of time from specialized physicians and nurses, in addition to a significant miss-rates of polyps by specialists. Automated polyp detection in colonoscopy videos has been demonstrated to be a promising way to handle this problem. {However, polyps detection is a challenging problem due to the availability of limited amount of training data and large appearance variations of polyps. To handle this problem, we propose a novel deep learning method Y-Net that consists of two encoder networks with a decoder network. Our proposed Y-Net method} relies on efficient use of pre-trained and un-trained models with novel sum-skip-concatenation operations. Each of the encoders are trained with encoder specific learning rate along the decoder. Compared with the previous methods employing hand-crafted features or 2-D/3-D convolutional neural network, our approach outperforms state-of-the-art methods for polyp detection with 7.3\% F1-score and 13\% recall improvement. 
\end{abstract}

\section{Introduction}
\label{sec:intro}
Colorectal cancer is a major cause of morbidity and mortality throughout the world \cite{haggar2009colorectal}. Early diagnosis is particularly relevant for colorectal cancer. The detection and removal of precancerous polyps in the colon may prevent later cancer. Screening the population for precursor lesions or manifest early colon cancer has been an important goal for decades \cite{meester2015public}\cite{simon2016colorectal}. Colonoscopy is often considered as gold standard because it allows an examination of the complete colon and it can remove pre-cancerous polyps immediately. It is a demanding procedure requiring significant amount of time from specialized physicians and nurses. Research works on colonoscopies have indicated significant miss-rates of 27\% for small adenomas (< 5 mm) and 6\% for adenomas of more than 10 mm diameter. {There are multiple factors that contribute to missed polyps at colonoscopy including quality of bowel preparation and experience of the colonoscopist \cite{bonnington2016surveillance}.} 

\begin{figure}
\centering
        \includegraphics[width=3cm,height=3cm]{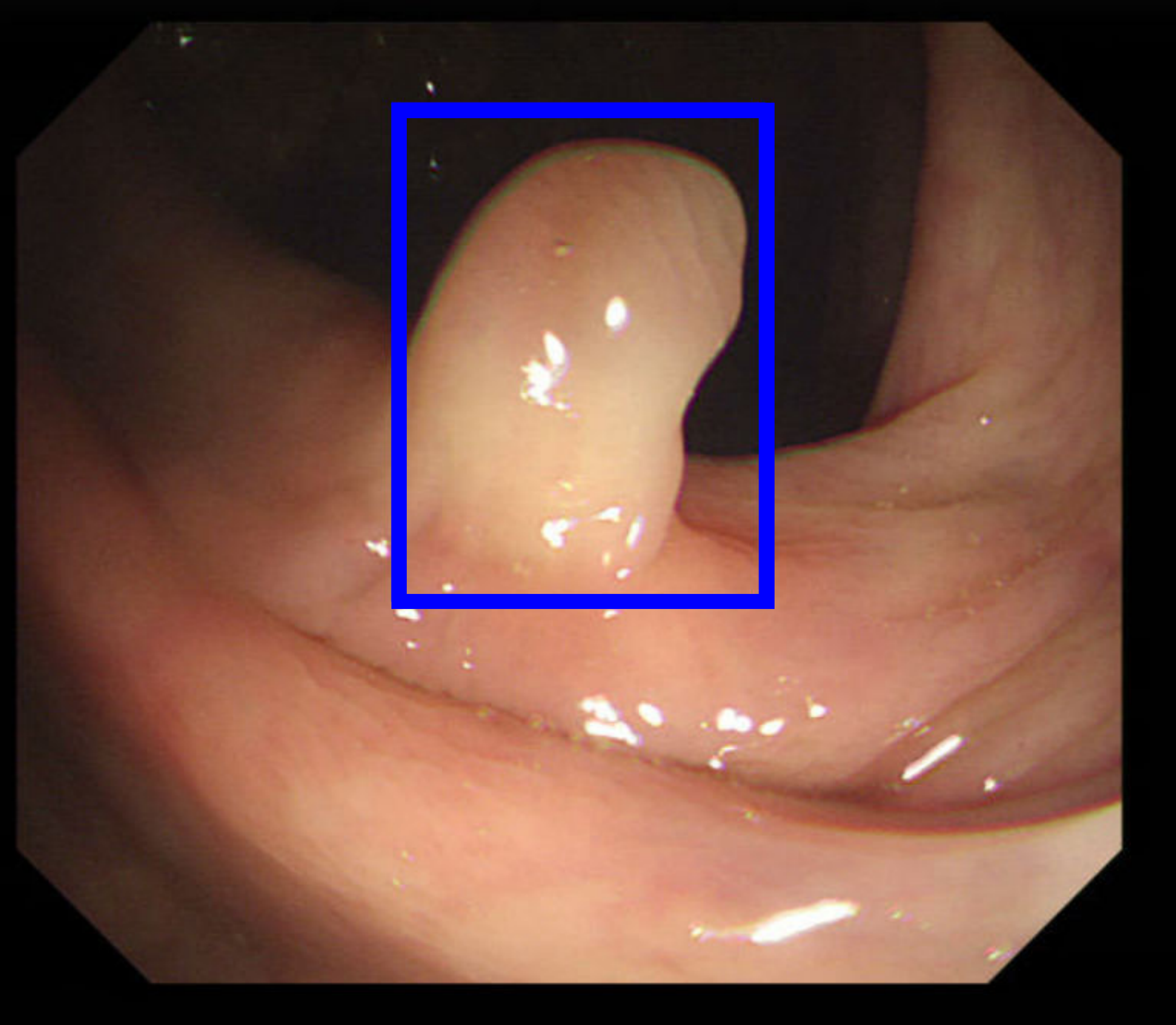}%
        \includegraphics[width=3cm,height=3cm]{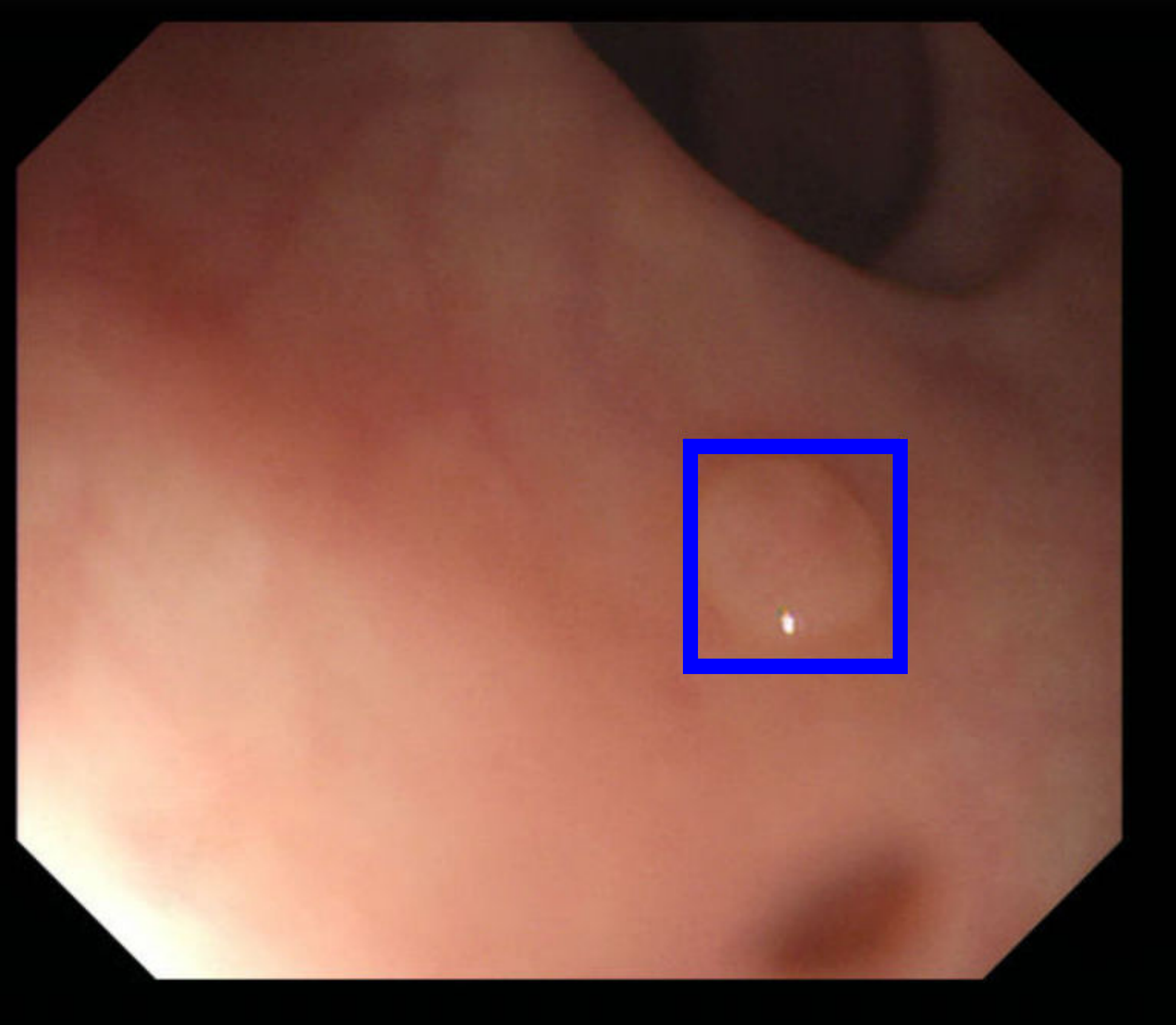}%
        \includegraphics[width=3cm,height=3cm]{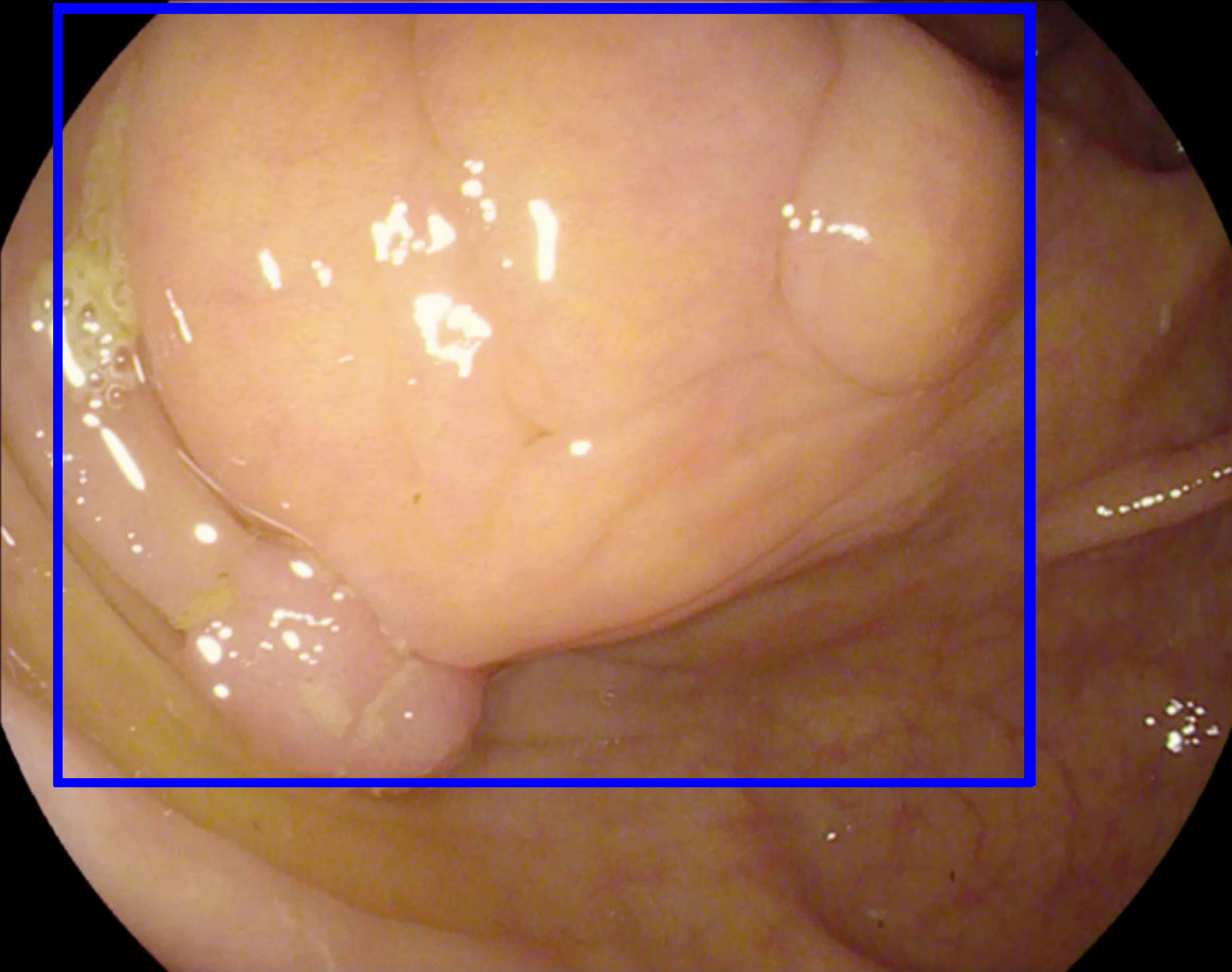}%
		\includegraphics[width=3cm,height=3cm]{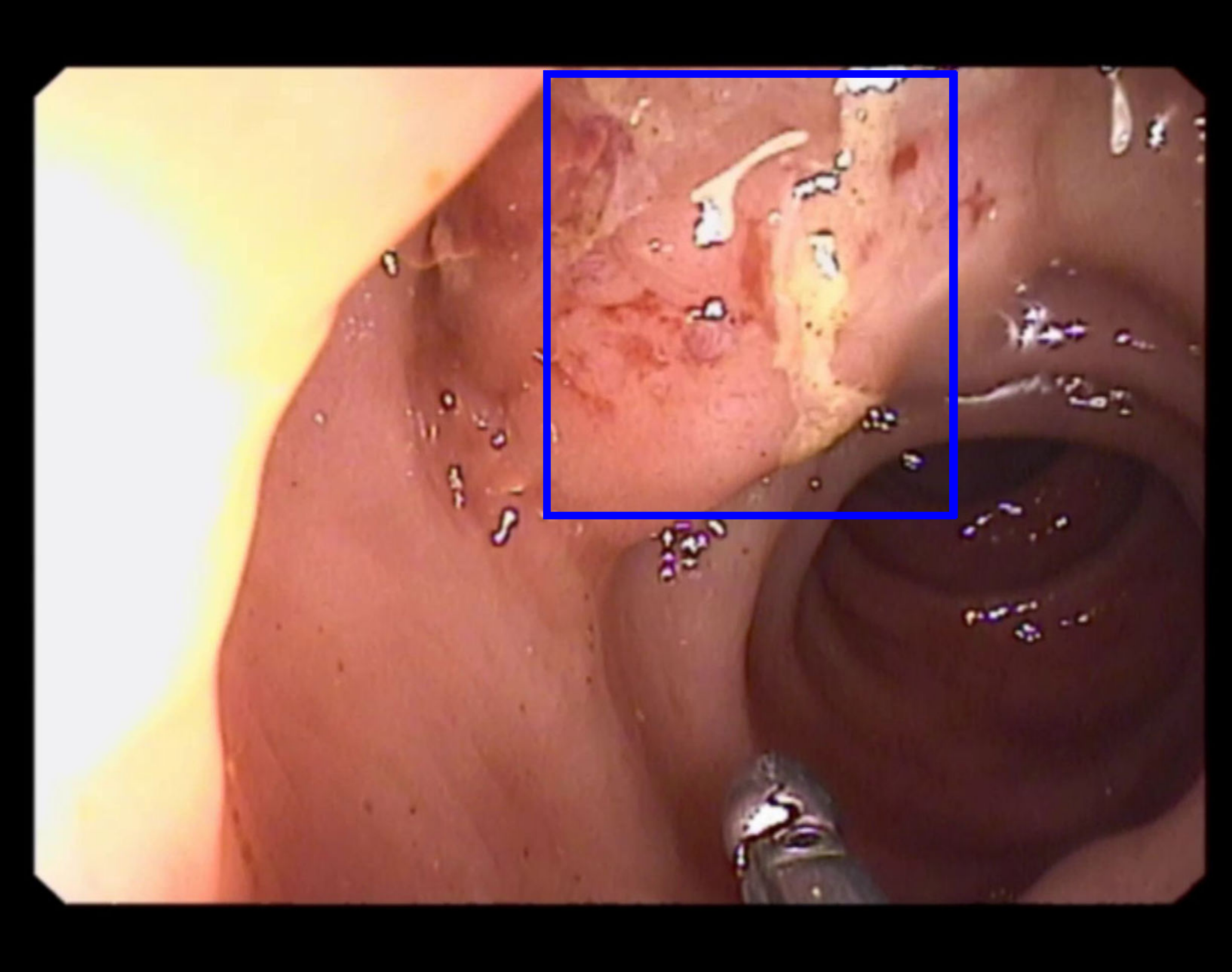}%
		\\%
		\includegraphics[width=3cm,height=3cm]{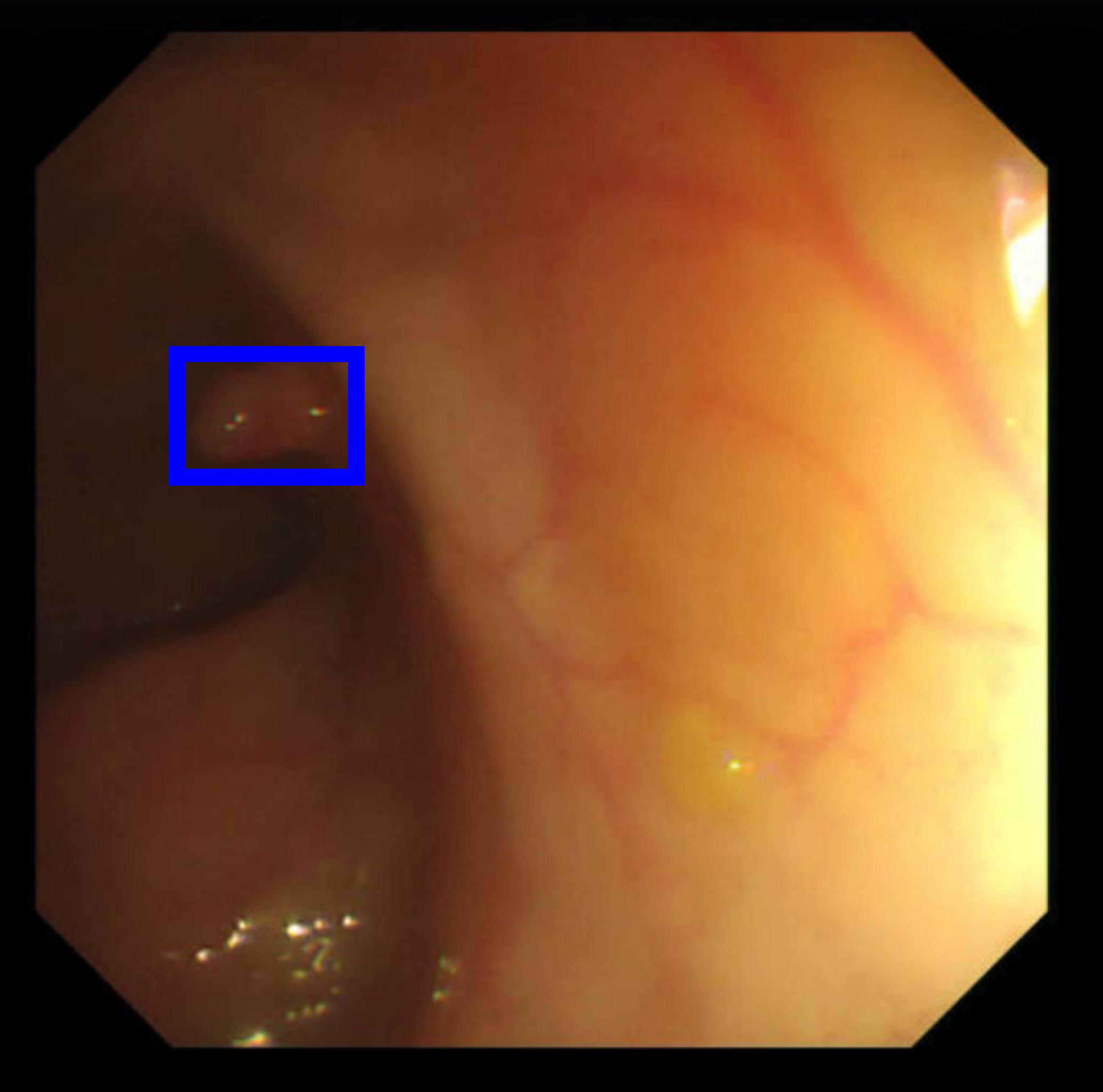}%
		\includegraphics[width=3cm,height=3cm]{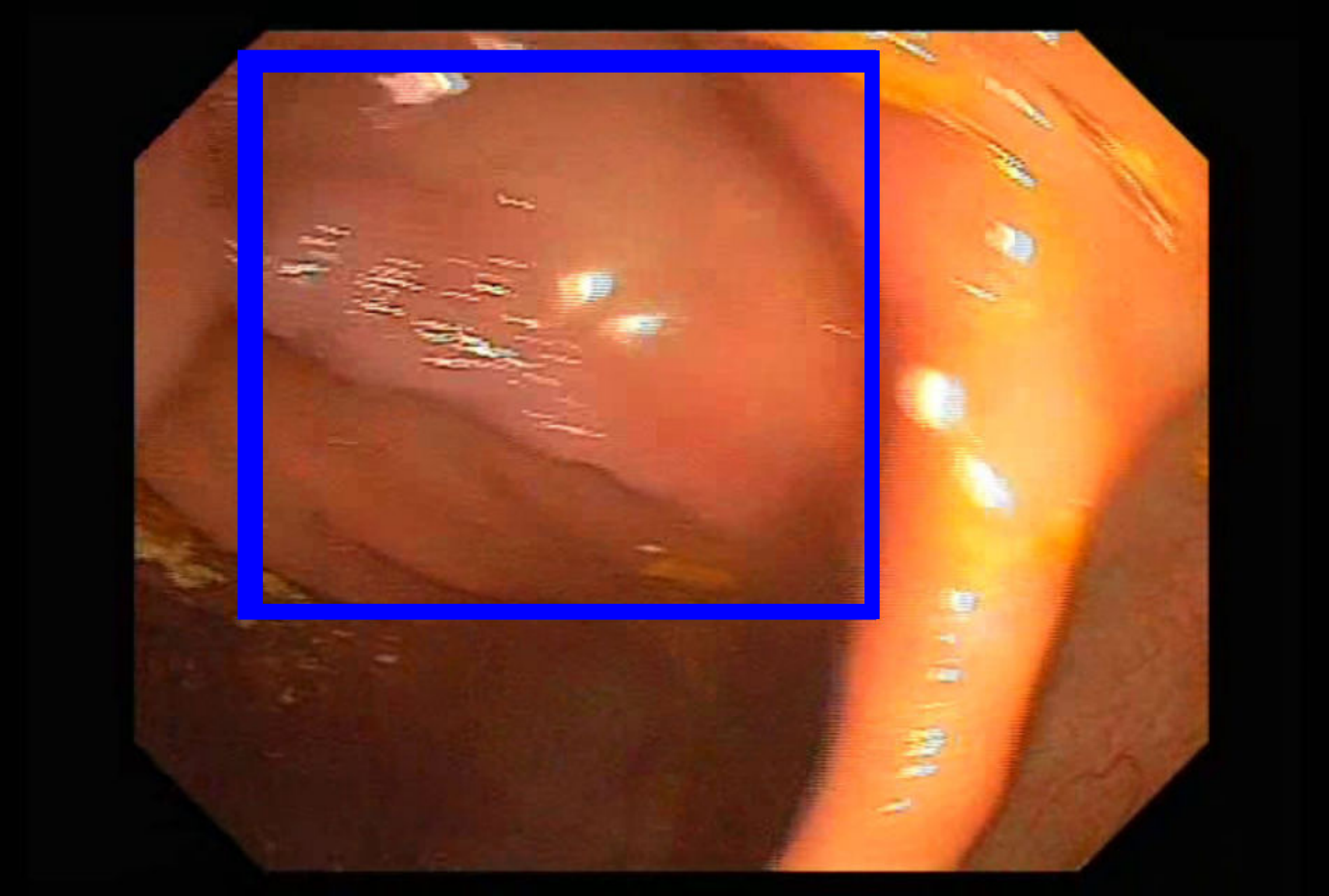}%
		\includegraphics[width=3cm,height=3cm]{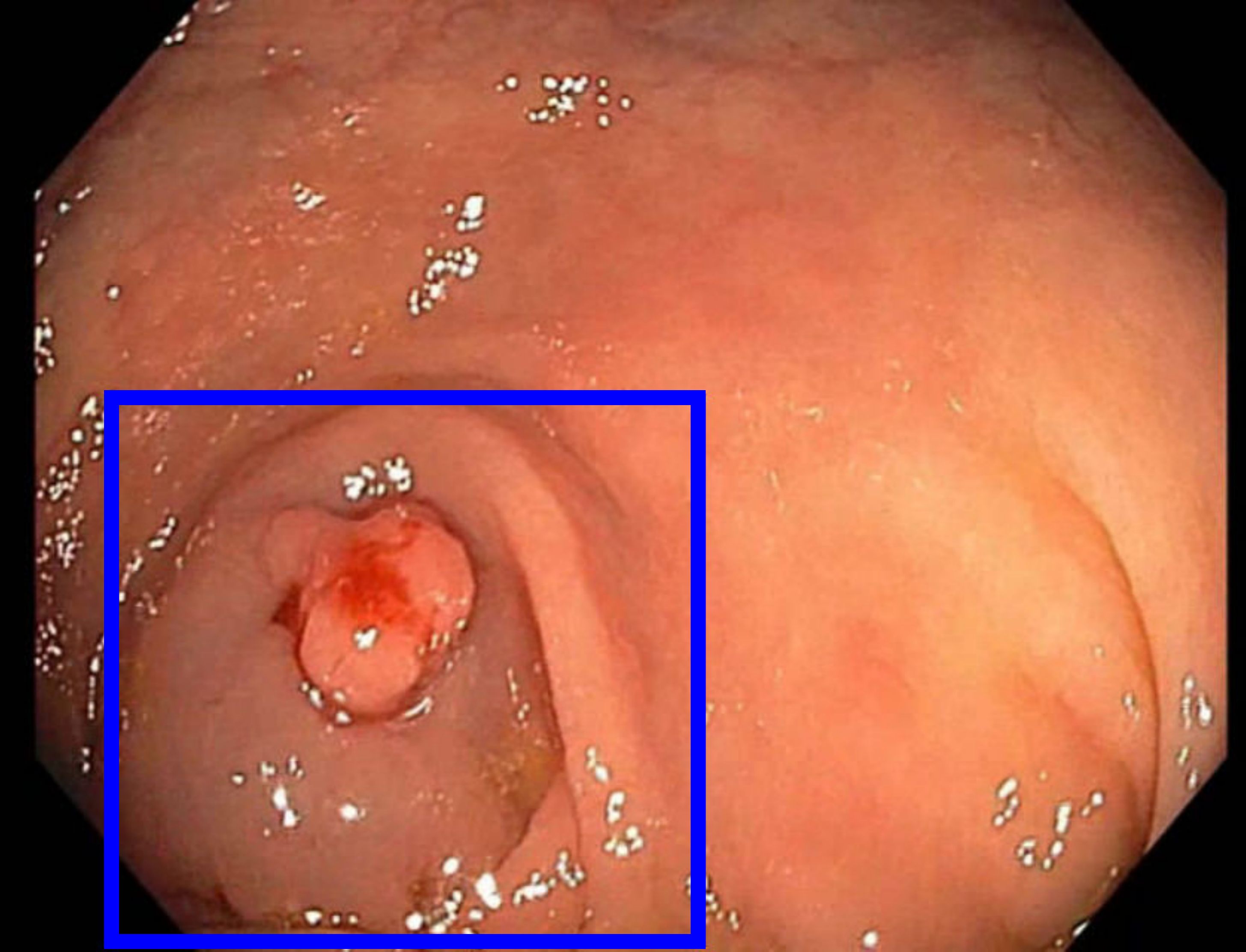}%
		\includegraphics[width=3cm,height=3cm]{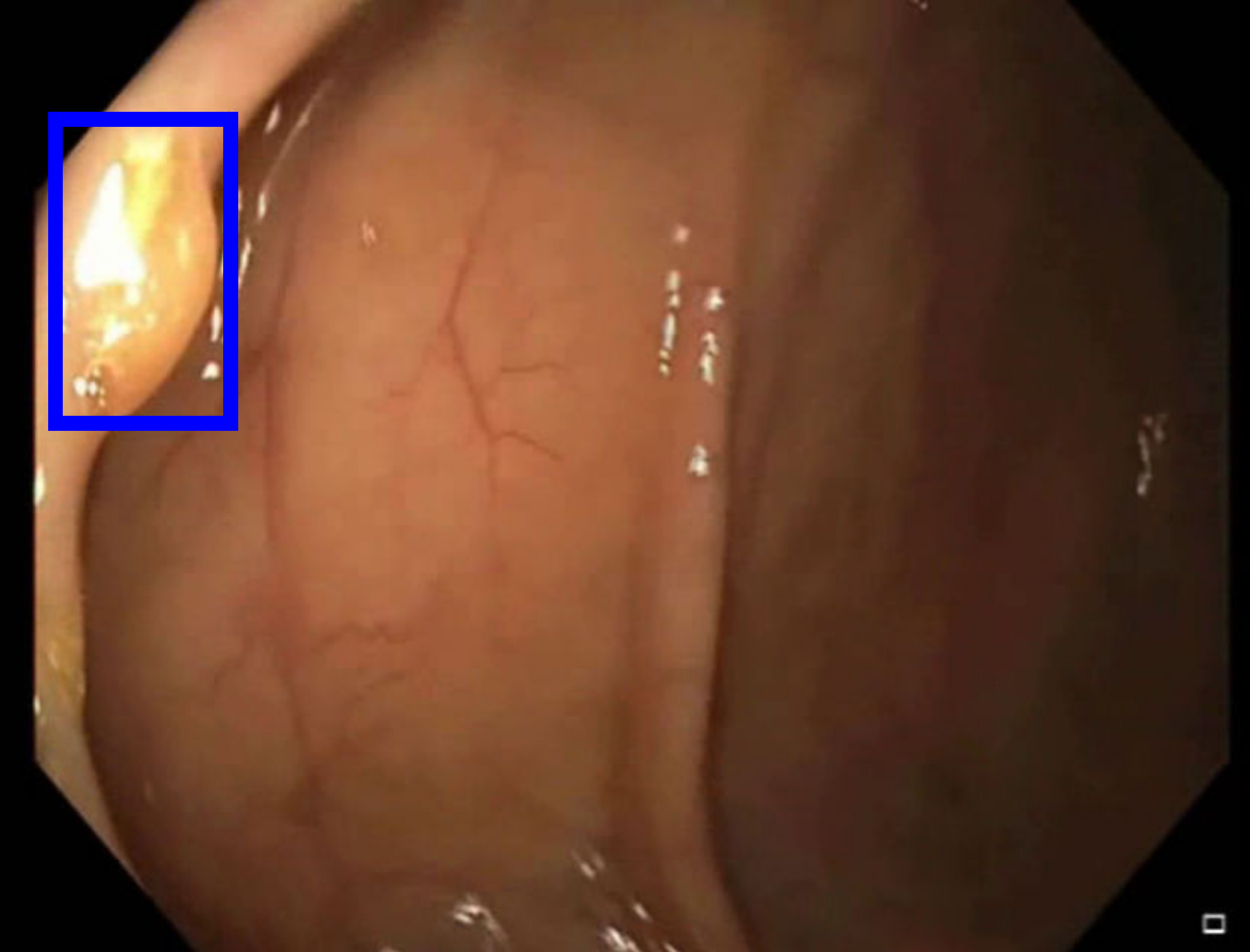}\\

	\caption{Appearance variability of polyps: Polyps vary in their shape, size and location within the large intestine. They may also have bleeding and abnormal patterns on surface.  }
	\label{fig:fig1}
\end{figure}

{Colon polyps grow in two different shapes. Pedunculated polyps have mushroom-like tissue growths and sessile polyps have flat structures that sits on the surface of the mucous membrane. Automated polyp detection techniques \cite{tajbakhsh2016automated}\cite{bernal2015wm}\cite{angermann2016active} can be used to mitigate the miss-rates and compliment colonoscopist in fast diagnosis. However, high precision and recall automated polyp detection is a challenging problem since colon polyps exhibit different size, orientation, color, and texture. We depicted these variations in Fig. \ref{fig:fig1}. }

{To detect colon polyps, researchers propose different methods \cite{tajbakhsh2016automated}\cite{bernal2015wm} based on hand-crafted features. These methods could not cope significantly with the problem of modeling distinctive features that differentiate polyps from normal mucosa. Therefore, deep convolutional neural networks (CNNs) are proposed to detect and localize colon polyps \cite{bernal2017comparative}\cite{yu2017integrating} that have presented very good performance. It is worth noting that CNNs based methods require large labeled datasets for training. However, large medical datasets are not readily available due to proprietary rights and privacy concerns.}

{To cope with the aforementioned problem, we propose Y-Net, a novel encoder-decoder hybrid deep neural network that can be trained with limited amount of training data. Our proposed Y-Net deep model is capable of detecting and localizing different colon polyps regardless of orientation, shapes, texture, and size. In fact, our method is inspired by U-Net \cite{ronneberger2015u}. Our method is different from U-Net in that we use multiple encoders with and without pre-trained network and learn a decoder network and without extensive data augmentation. Y-Net model aims to learn decoder network from scratch while fine tuning encoder networks with different learning rates. Our method is conceptually simple, relying on the pretrained VGG network \cite{simonyan2014very} as one of the encoders and a matched decoder network with the novel introduction of sum-skip-concatenation based connections to allow a much deeper network architecture for a more accurate segmentation. The key difference with the existing models is that we introduced a pre-trained encoder network that is augmented with untrained mirrored network and a decoder network that uses discriminative cost function to localize and detect polyps.}

Our main contributions are: (1) A novel encoder-decoder network that uses a pretrained model as one of its encoder with mirrored untrained network and a decoder that is trained from scratch, making it more practical to train for polyp detection with limited amount of data. (2) A sum-skip-concatenation connection to a decoder network that allows re-use of pretrained encoder network weights. (3) Qualitative and quantitative experiments on  MICCAI 2015 challenge on polyp detection \cite{bernal2017comparative}\cite{bernal2015wm}\cite{tajbakhsh2015} and comparison with state-of-the-art works \cite{yu2017integrating}, show that our method outperforms in polyp detection with 7.3\% F1-score and 13\% recall improvement.

\section{Previous Work}
\label{sec:background}
There is a large literature dedicated to the topic of deep learning for image segmentation and detection. Our review here is brief due to space limitations, and is intended to highlight the broad approaches of existing polyp detection algorithms and to provide appropriate background for our work. For a current state-of-the-art and thorough review of deep learning for image segmentation and detection, please refer to \cite{garcia2017review}. Mainly for polyp detection, two approaches are seen in literature: hand-crafted features with classifiers and deep-learning. \newline
{\bf Hand-crafted features:\quad} CVC-CLINIC \cite{bernal2015wm} proposed polyp localization algorithm by modeling the appearance of the polyps. Even though there are flat surface polyps but they assume all the polyps have protruding surfaces. Consequently, detection of the polyp is done through intensity valleys detection. They introduce a pre-processing step which helps to filter out other valley-rich structures like blood vessels. Similarly, Silva et al. \cite{silva2014toward} introduced a method which is influenced by the psycho-visual methodology of clinicians. Initially geometric features of polyps are exploited to get a region of interest (ROI) through the Hough transform. In the 2nd stage, texture features are used in an ad-hoc boosting-based classifier to filter out the ROI which does not contain any polyps. Generally, handcrafted features provide less correct detections and significantly more false alarms \cite{bernal2017comparative}. \newline
{\bf Deep learning approach:\quad}{Different methods based on deep learning models are proposed  \cite{bernal2017comparative}\cite{billah2017automatic}\cite{tajbakhsh2015}\cite{yu2017integrating}.} Tajbakhsh et al. \cite{tajbakhsh2015} came up with a two level approach for automatic polyp detection. Initially, geometric features like shape and size of polyps are exploited to get a set of candidate regions of having polyps. In the 2nd level, an ensemble of CNN are used for the classification. The output of the CNNs are averaged to get a probabilistic map for the existence of polyps in the frame. Yu et al. \cite{yu2017integrating} proposed offline and online three-dimensional  deep learning integration framework. With their approach they used online and offline 3-D representation learning to reduce the number of false positives and further improve the discrimination capability of the network for a specific video. Other teams on the MICCAI 2015 challenge \cite{bernal2017comparative} such as CUMED used VGG type architecture with down-sampling path and up-sampling path with auxiliary classifiers to alleviate  the problem of vanishing gradients and encourage the backpropagation of gradient flow. Similarly, OUS used pretrained AlexNet model and CaffeNet, modifying the input size to 96 x 96  and applying sliding window on three scale of the original input image. 

{Our proposed Y-Net method is also based on deep learning model and is inspired by U-Net \cite{ronneberger2015u}, a fully convolutional network. They used excessive data augmentation by applying elastic deformations to the available training images. However, our method explore multiple encoders with and without pre-trained network and learn a decoder network and without extensive data augmentation.} 

\section{Methodology}
Our overall approach is based on using robust pre-trained and un-trained encoder networks. The framework (illustrated in Fig. \ref{fig:teaser}) {consists of two fully convolution encoder networks which are connected to a single decoder network. The main goal for having two encoders network is to address the performance loss due to domain-shift from pre-trained network (natural images) to testing (polyp data), leading to degradation in performance. For example, a pre-trained models trained on natural images do not generalize well when applied to medical images. It is assumed that fine-tuning a pre-trained network works the best when the source and target tasks have high degree of similarity.  Therefore, our approach focuses using the pre-trained model features optimally by slow fine-tuning the pre-trained network and aggressive learning on the second encoder for a better generalization on the test set.  In the next sub-sections, we describe each of the network components, and then the losses used to train the network. 

\subsection{Network Architecture}
The architecture of our model is shown in Fig. \ref{fig:teaser}. It consists of two contracting paths on the left, i.e. encoders, and expanding path to the right, i.e. a decoder, that matches the input dimension. The decoder outputs a binary mask segmentation of the polyp on the last layer. 
\newline
{\bf Encoder one:\quad}It follows the typical architecture of VGG19 network \cite{simonyan2014very}, which has been widely used as the base network in many vision application. This encoder uses a pre-trained weights of VGG19 trained on ImageNet dataset. The last fully connected layer of the network that was trained on 1000 ImageNet classes is truncated.  Usage of these pre-trained models allows us in that the pre-trained model will already have learned features that are relevant to our own classification problem such as edges, curves and etc. \newline
{\bf Encoder two:\quad}It follows the same VGG19 network architecture without the fully connected layer. The same input image is given to both encoders. It has 16  convolutional layers with $3 \times 3$ convolutions and $2 \times 2$ max-pooling layer. The weights of the network are initialized using Xavier normal initializer \cite{glorot2010understanding}. Moreover, it uses SELU activation function \cite{klambauer2017self} to improve the back propagation as well as mitigates the vanishing and exploding gradients problem \cite{DBLP:journals/corr/abs-1712-05577}.  \newline
{\bf Decoder:\quad}
\begin{figure}[t!]
	\centering
	\includegraphics[width=0.7\linewidth]{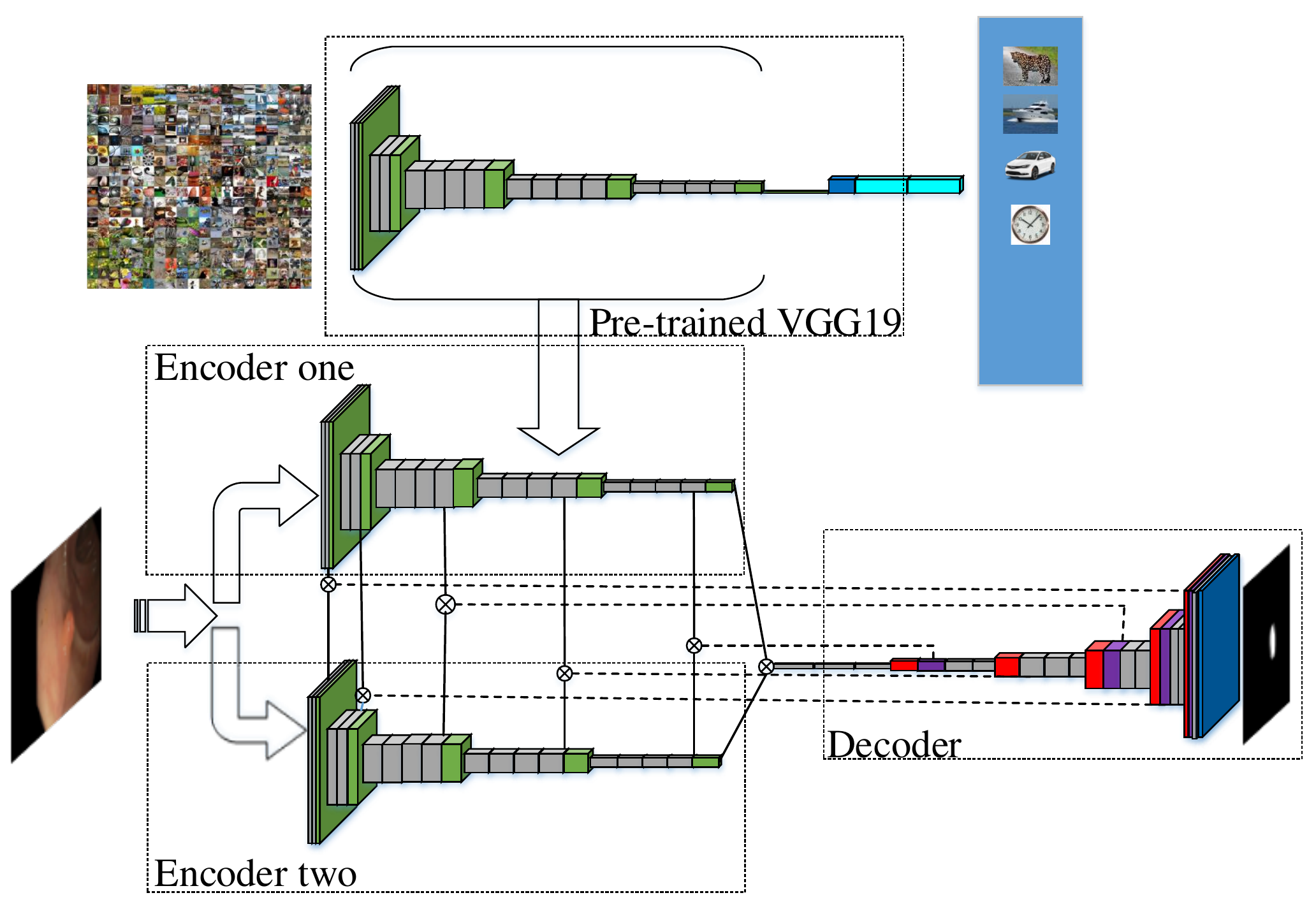}
	\begin{tabular}{rrrrrrr}
	
		\includegraphics[width=0.3cm]{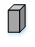}Conv&
		\includegraphics[width=0.3cm]{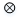}Sum&
		\includegraphics[width=0.3cm]{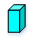}Fully Connected&
		\includegraphics[width=0.3cm]{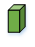}Max-pooling&
		\includegraphics[width=0.3cm]{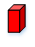}Up-sampling&
		\includegraphics[width=0.3cm]{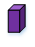}Concat-Conv\\
		
	\end{tabular}
	\vspace{0.1cm}
	\caption{Y-Net: proposed architecture. The top row shows VGG19 network pre-trained on ImageNet dataset with 1000 classes such as cars, tiger and etc. The weights are transfered to encoder one network as shown above. Given an input image, it is fed to both encoders. The weights of last convolution at each depth of the encoder are summed and concatenated to the same spatial depth of the decoder. The output of the decoder is a binary label for foreground (polyp) and background (non-polyp).}
	\label{fig:teaser}
\end{figure}
The decoder network consists of five upsampling blocks and one final convolution block with a filter size of $1 \times 1$. Each upsampling block has the structure of Upsampling-concatenation followed by three blocks of CONV-SELU-BatchNorm(BN), except for the final layer which uses a $1 \times 1 $ convolution with sigmoid for generating the final output mask. Compared with the other encoder-decoder architectures such as U-Net \cite{ronneberger2015u}, our decoder is different in: (1) The decoder is not architecturally symmetric with the encoders. (2) The decoder is much deeper than the encoder. This design is due to the fact that with the limited polyp training data, a deeper decoder network would learn features from each scale of the encoder inputs that are concatenated with the same-scale decoder layer. 
\newline
{\bf Sum-skip-concatenation:\quad}Let $d$ be the depth of encoder network $k$, $X^{i,j}_k$, with $j$ index of convolution $X$ at depth $i$. Let $\mathcal{U(\cdot)}$ be up sampling operation. For $j_{th}$ convolution in $i_{th}$ depth of $k_{th}$ encoder network $X^{i,j}_k$, the response of encoder convolution at depth $X^i$ is estimated as follows:
\begin{equation}
\label{eq:1}
X^{i}=\sum_{k=1}^{2}X^{i,j}_k, j=2,  i \leq 1 \leq  2, j=4,  i \leq 3 \leq  5
\end{equation}
For each depth of the decoder, the first feature input is computed by concatenating up-sampled $\mathcal{U}(d-1)$ response of the decoder with the skipped summed output of the encoder. This connection has many advantages for training: (1) It can combine easily a pre-trained and un-trained network (Eq. (\ref{eq:1}))  with skip connection to a decoder. (2) by using different learning rate for both encoders, the un-trained encoder is trained aggressively without losing the learned pre-trained weights in training the decoder.  
\newline

\subsection{Model Learning and Implementation}
Given the network architecture outlined above with one of the encoder pre-loaded with pre-trained VGG19 weights, we explain next the optimization objectives and training strategy.  
\newline
{\bf Loss function:\quad}The output layer in the decoder consists of a single plane for foreground detected polyp. We applied convolution with sigmoid activation to form the loss. Let $P$ and $G$ be the set of predicted and ground truth binary labels, respectively. The weighted binary cross-entropy and dice coefficient loss between two binary images is defined as:
\begin{equation}
\label{eq:2}
 \mathcal{L}(P,G)= -\frac{1}{N} \sum_{i=1}^{N}\Big(\frac{\lambda}{2}  \cdot g_i \cdot \log p_i\Big) +\Bigg(1- \frac{ 2\sum_{i=1}^{N}(g_i \cdot p_i)+\epsilon}{\sum_{i=1}^{N}(p_i)+\sum_{i=1}^{N}(g_i)+\epsilon}\Bigg)
\end{equation}
where $\lambda$ and $\epsilon$ are false negatives (FN) penalty and smoothing factor, respectively.  In order to penalize FN more than false positives (FP) in training our network for highly imbalanced data, the first term in Eq. (\ref{eq:2}) penalizes FN and the second term   weighs FPs and FNs (precision and recall) equally. In other words, the second term is the same as the negative of F1-score. This is to avoid miss detection of polyps, as it is more critical to miss a polyp than giving a FP. Hence, the summed loss function gives a good balance between FN and FP. 
\newline
{\bf Learning rates:\quad}Since the pre-trained encoder weights are initialized with VGG19, they are good, compared to randomly initialized weights, in extracting basic image features such as edges and curves.  Therefore, it would be beneficial while training not to distort them too much. Hence, we propose encoder specific adaptive learning rates. The parameter update equation for  RMSProp (Root Mean Square Propagation) gradient descent becomes 
\begin{equation}
\label{eq:3}
\theta_{t+1}=\theta_{t}- \frac{c \cdot \eta}{\sqrt{E[g^2]+\epsilon}} \cdot g_t
\end{equation}
where $c=0.01$ for pre-trained encoder and $c=1$ for encoder two and decoder, $\theta$ is a model parameter with a learning rate $\eta$ and $E[g^2]$ is the running average of squared gradients. In this way, encoder two is learned aggressively while fine-tunning the pre-trained encoder.
\section{Experiments}
{\bf Dataset:\quad}Experiments are conducted on the ASU-Mayo clinic polyp database \cite{tajbakhsh2016automated} of MICCAI 2015 Challenge on polyp detection. The dataset consists of 20  and 18 short segment colonoscopy videos for training and testing with pixel level annotated polyp masks in each frames respectively. The dataset contains colonscopy videos taken under variations in procedures (i.e. a careful colon examination while others show a fast colon inspection), has maximum variations in colonoscopy findings such as polyp variations, different resolutions, and existence of biopsy instruments. Among 20 training videos, ten videos have polyps inside and the other ten videos have no polyp. There are total 4278 frames with polyps in the training set and 4300 frames in test set. 
\newline
{\bf Data augmentation:\quad}To increase robustness and reduce overfitting on our model, we increase the amount of training data. First, frames with polyp are doubled pre-training by applying random rotation (\ang{10} to \ang{350}), zoom (1 to 1.3), translation in x,y (-10 to 10) and shear (-25 to 25) followed by centering the polyp and cropping padded regions. Second, during training, for each frame with polyp, random cropping of the non-polyp region as well as perspective transform is applied with probability of 0.3 and 0.4 with random horizontal and vertical flip respectively. We also tried applying contrast enhancement methods such as CLAHE (Contrast-limited adaptive histogram equalization) \cite{zuiderveld1994contrast} and gamma correction, but that did not improve the detection accuracy.  
\newline
{\bf Implementation Details:\quad}Our model is implemented on Tensorflow and Keras library with a single NVIDIA GeForce GTX 1080 GPU. Due to the different image sizes in the dataset, we first crop large boundary margins and resize all images into fixed dimensions with spatial size of $224 \times 224$ before feeding to both encoders and finally normalized to $[0,1]$. We use custom RMSProp (Eq. (\ref{eq:3})) as the optimizer with batch size 3 and learning rate $\eta$ set to $0.0001$. We monitor dice coefficient and use \textit{early-stop} criteria on the validation set error. 
\newline
{\bf Evaluation Metrics:\quad}We evaluate the effectiveness of our model using recall and precision rate. If recall or true positive rate is low, then the model will miss finding polyps which can lead to late stage diagnosis for colorectal cancer. If precision is low, then it will add further examination and work for the gastroenterologist. Moreover, negative classes out number the positives with large margin, i.e. there are more frames without polyp than with polyp \cite{berhane2009incidental}. Hence, we employed F1-score and F2-score as they give a balance between missed polyps and false alarms. In order to calculate F1-score and F2-score we use the following equations:
\begin{equation}
\label{eq:4}
F1=\frac{2PR}{P+R}, F2=\frac{5PR}{4P+R} \quad
\text{and}\quad
P=\frac{N_{TP}}{N_{TP}+N_{FP}}, R=\frac{N_{TP}}{N_{TP}+N_{FN}}, 
\end{equation}
where $N_{TP}$ and $N_{TN}$ are the number of true positives and negative and $N_{FP}$ and $N_{FN}$ are the number of false positive and negative, respectively. The detection is considered to be correct when the intersection over union (IoU) between detection
box $P$ and ground-truth bounding box  $G$ is greater than zero, otherwise the detection box is assumed as a false positive. Each pixel in the detection mask is considered as true detection if is above $0.9$. IoU is formulated as:
\begin{equation}
\label{eq:5}
IoU=\frac{P \cdot G}{\sum{P}+\sum{G}+\epsilon} *100\%
\end{equation}
\begin{figure}
	\centering
	\includegraphics[width=2cm,height=2cm]{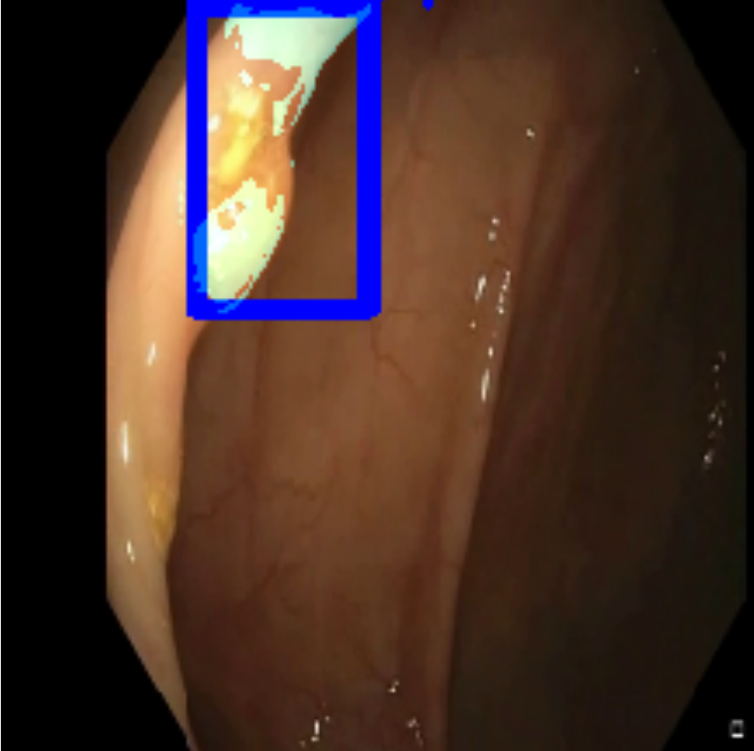}%
	\includegraphics[width=2cm,height=2cm]{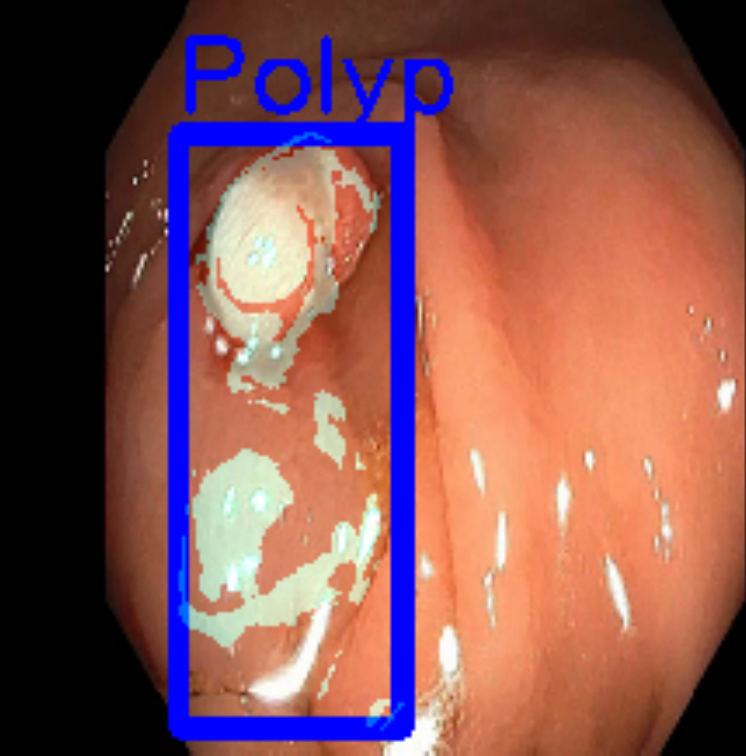}%
	\includegraphics[width=2cm,height=2cm]{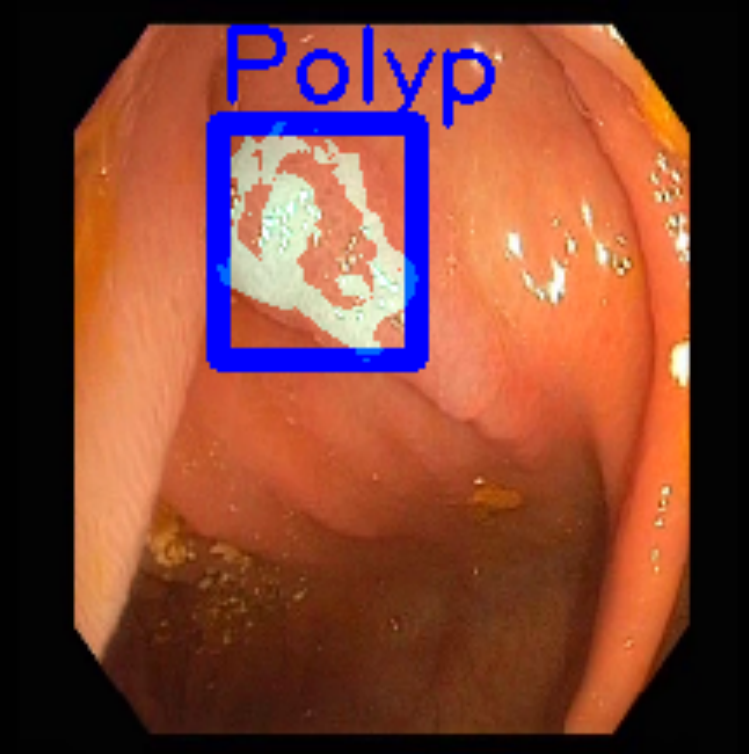}%
	\includegraphics[width=2cm,height=2cm]{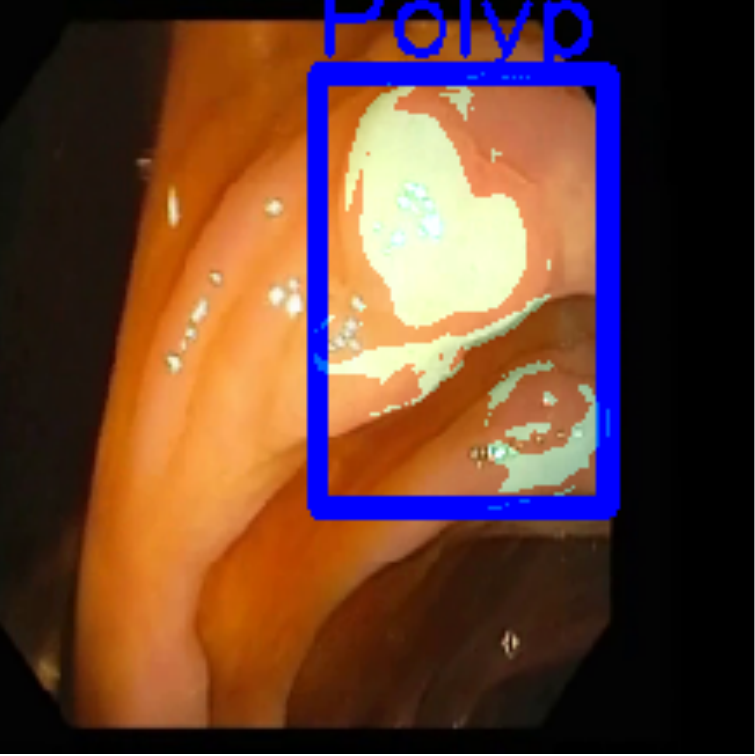}%
	\includegraphics[width=2cm,height=2cm]{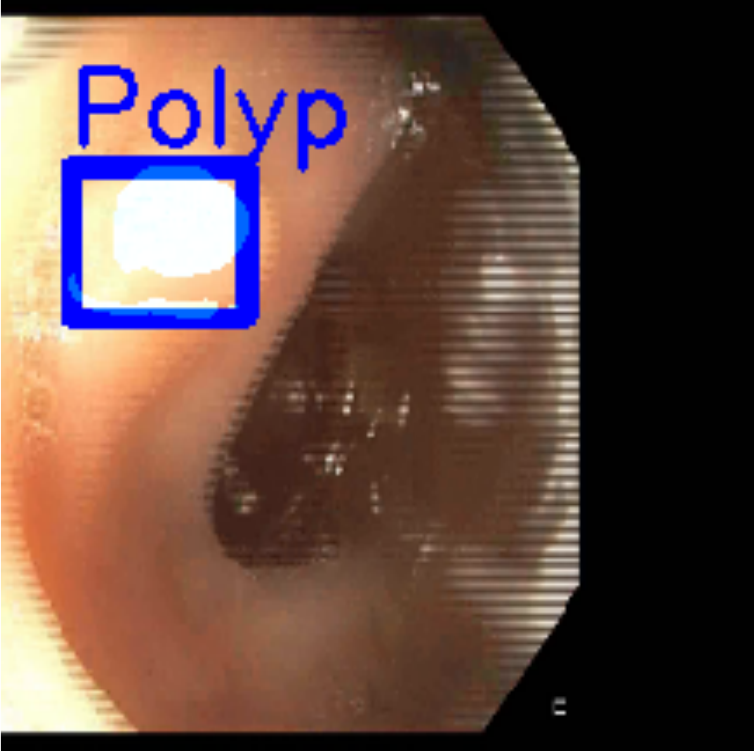}%
	\includegraphics[width=2cm,height=2cm]{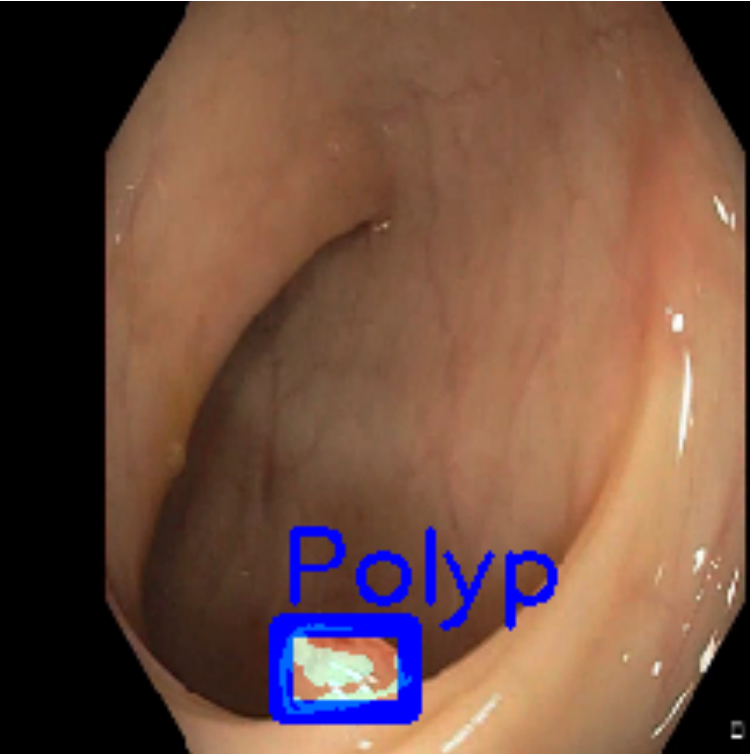}%
	\\%
	
	\includegraphics[width=2cm,height=2cm]{images/10_190.pdf}%
	\includegraphics[width=2cm,height=2cm]{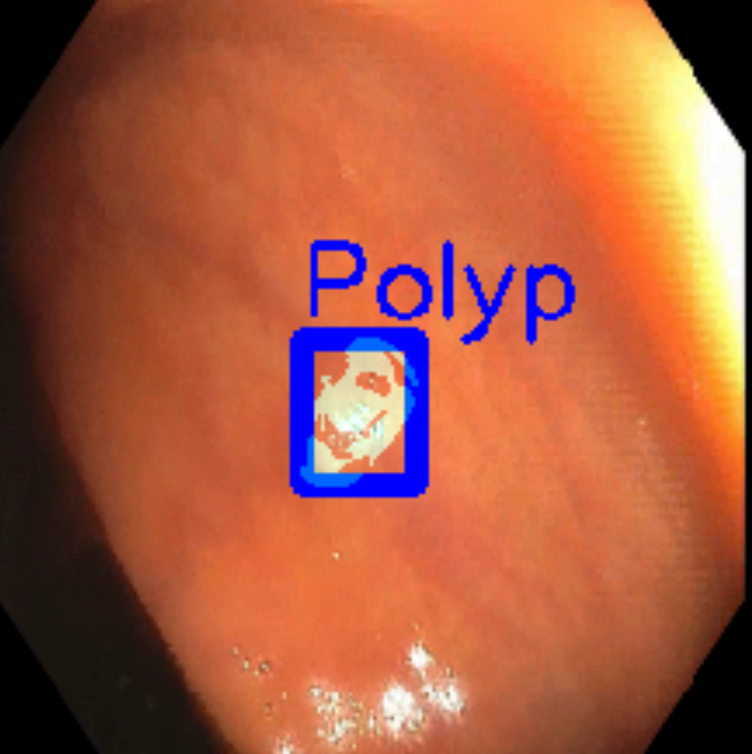}%
	\includegraphics[width=2cm,height=2cm]{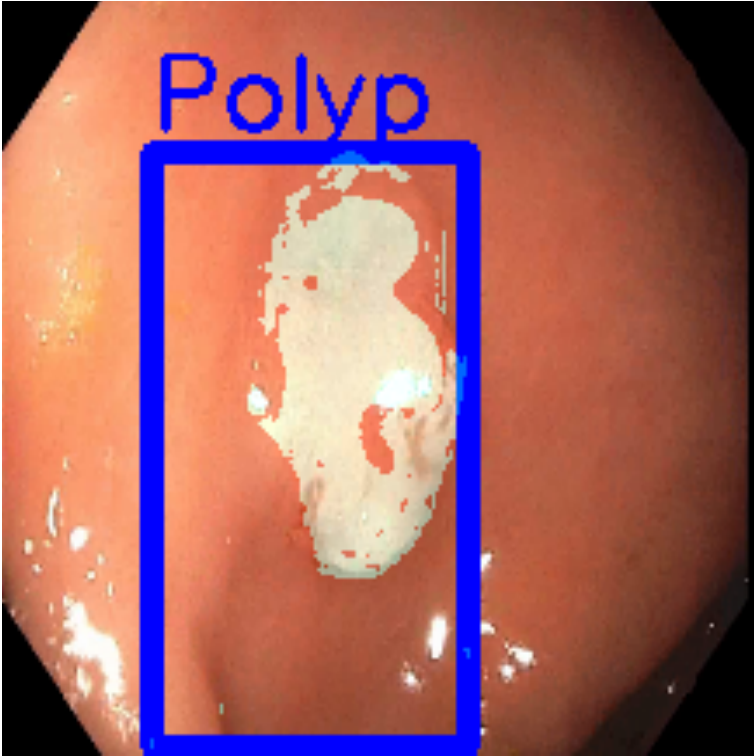}%
	\includegraphics[width=2cm,height=2cm]{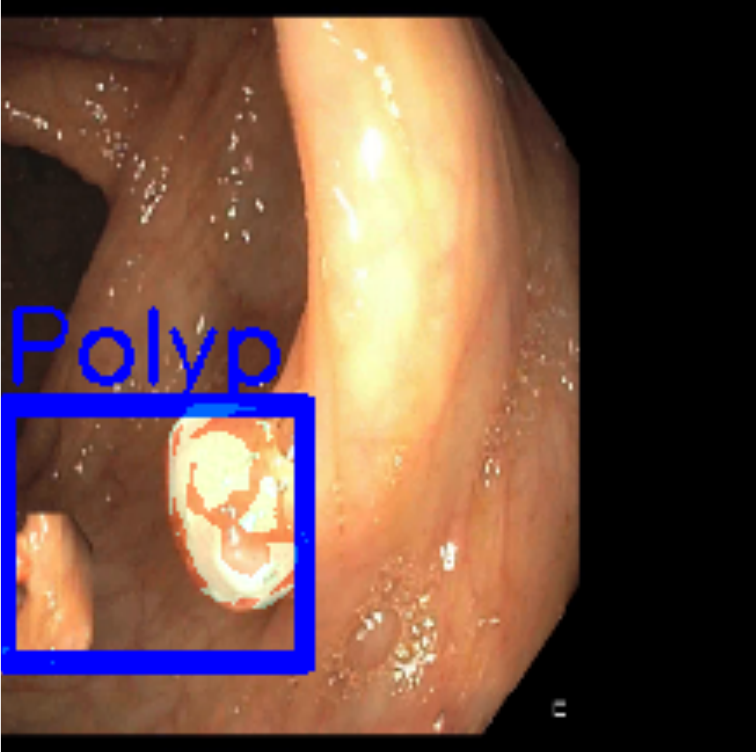}%
	\includegraphics[width=2cm,height=2cm]{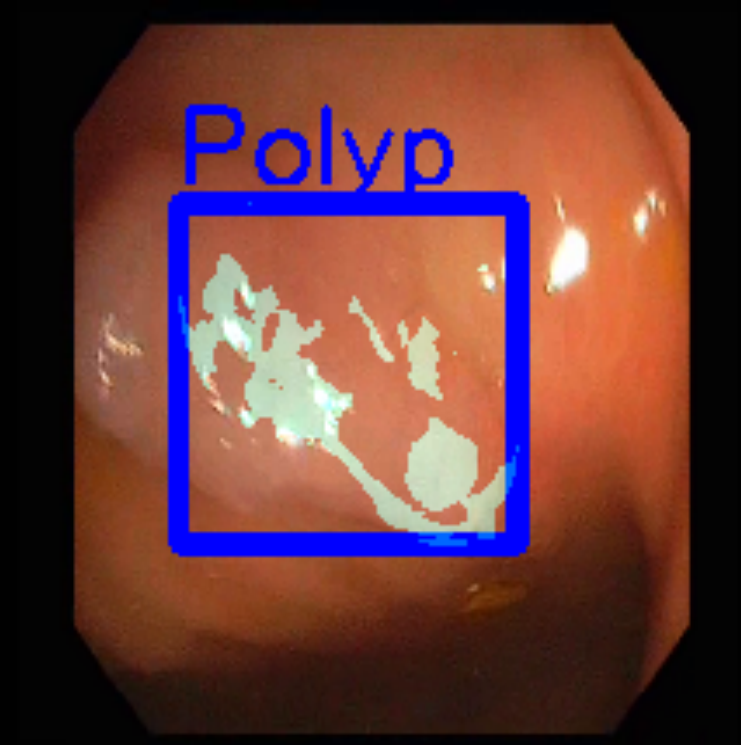}%
	\includegraphics[width=2cm,height=2cm]{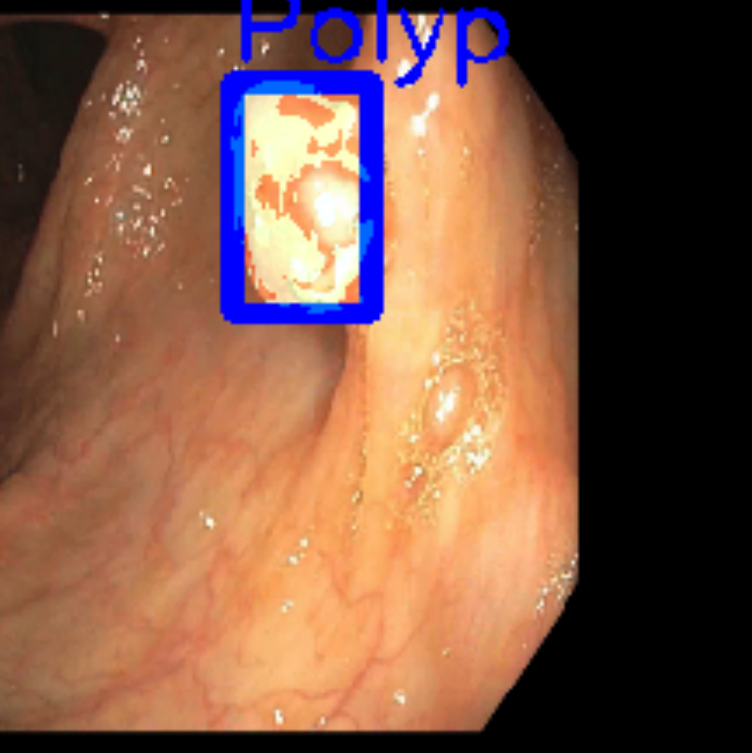}%
	\\%
	\includegraphics[width=2cm,height=2cm]{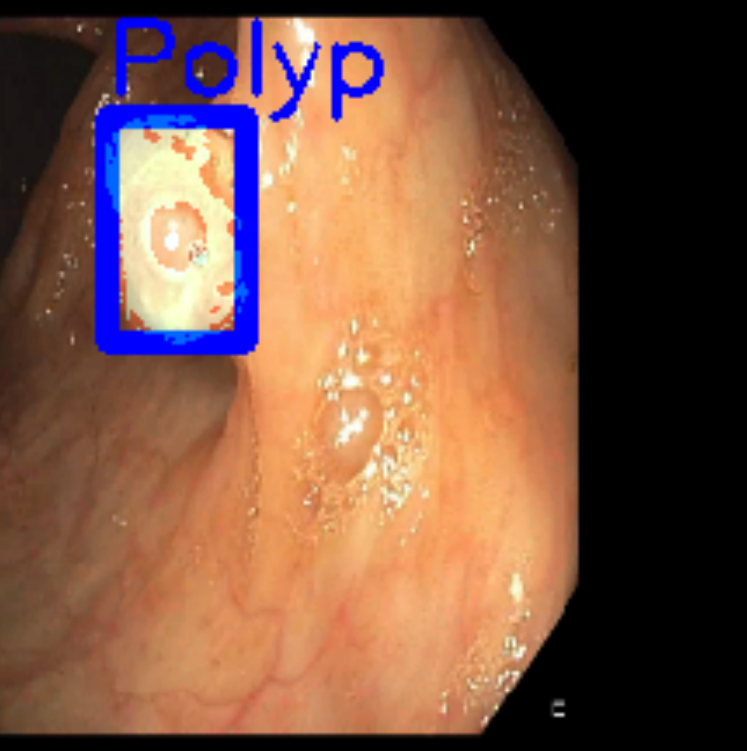}%
	\includegraphics[width=2cm,height=2cm]{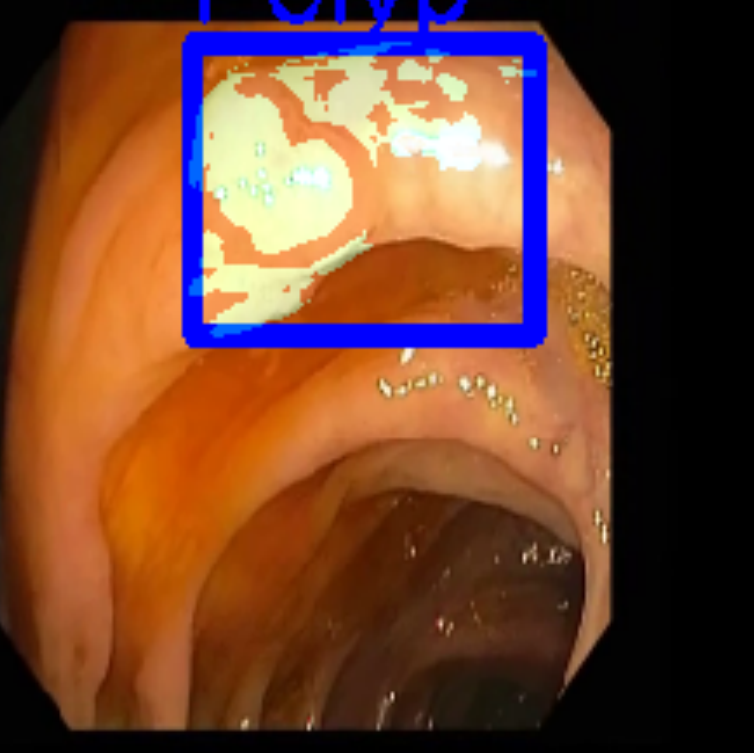}%
	\includegraphics[width=2cm,height=2cm]{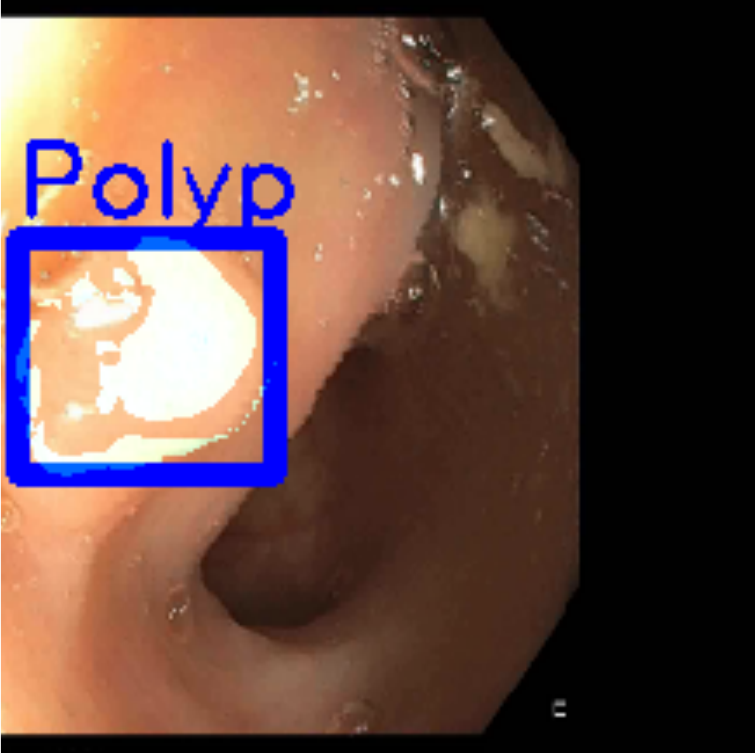}%
	\includegraphics[width=2cm,height=2cm]{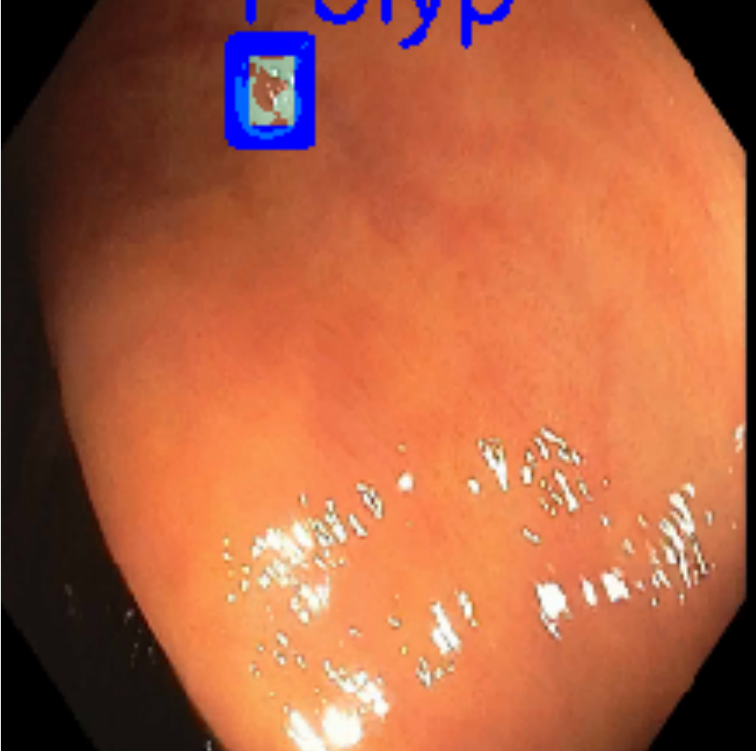}%
	\includegraphics[width=2cm,height=2cm]{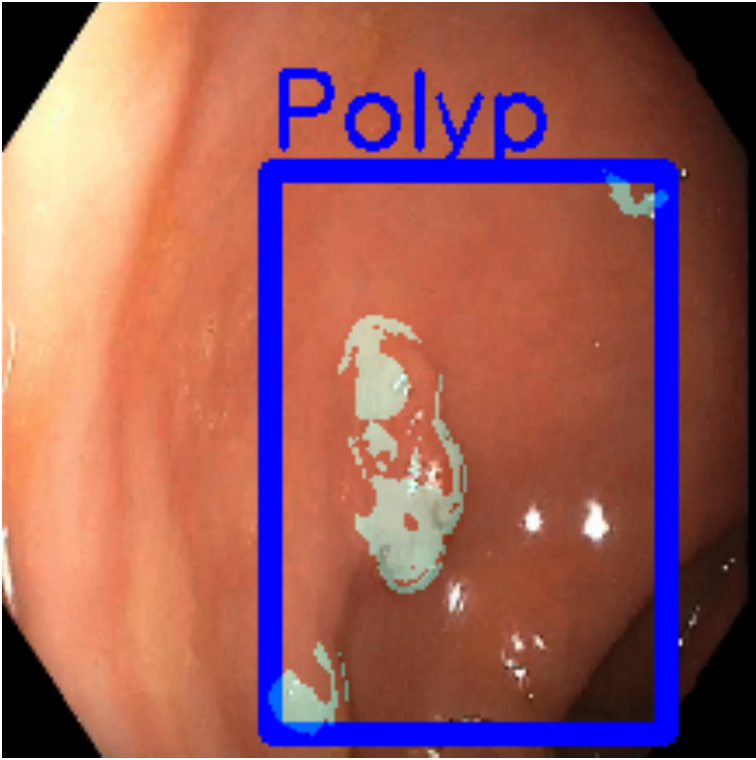}%
	\includegraphics[width=2cm,height=2cm]{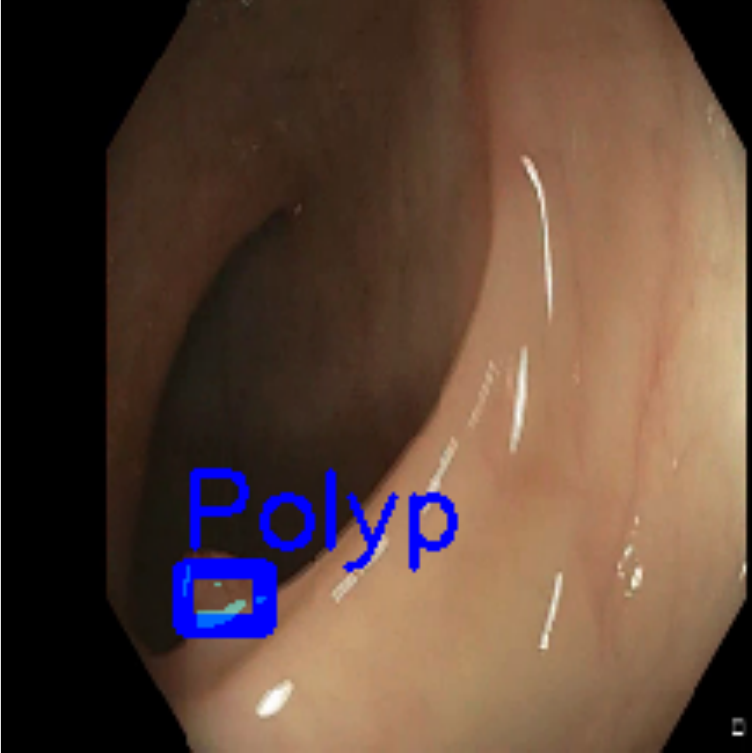}%
	\\
	
	\includegraphics[width=2cm,height=2cm]{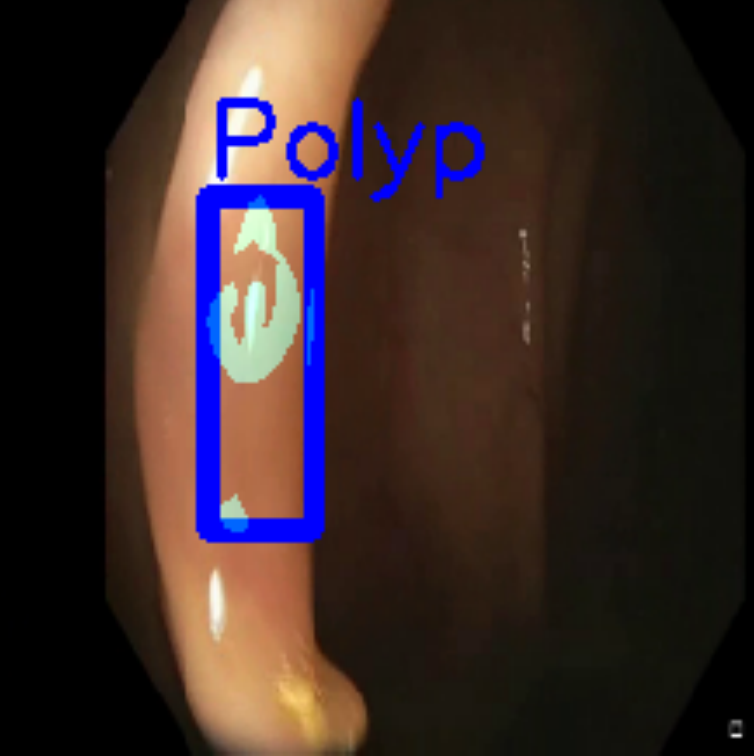}%
	\includegraphics[width=2cm,height=2cm]{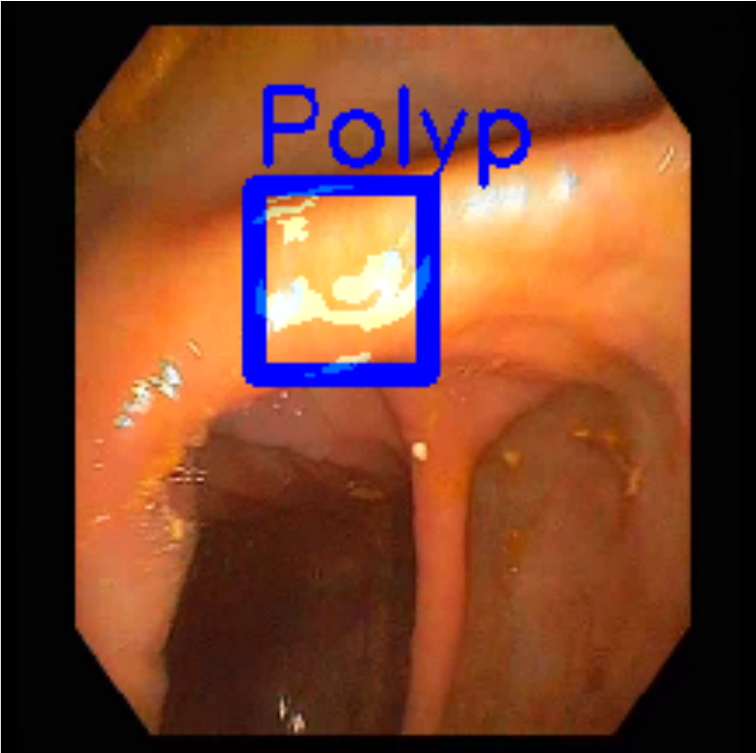}%
	\includegraphics[width=2cm,height=2cm]{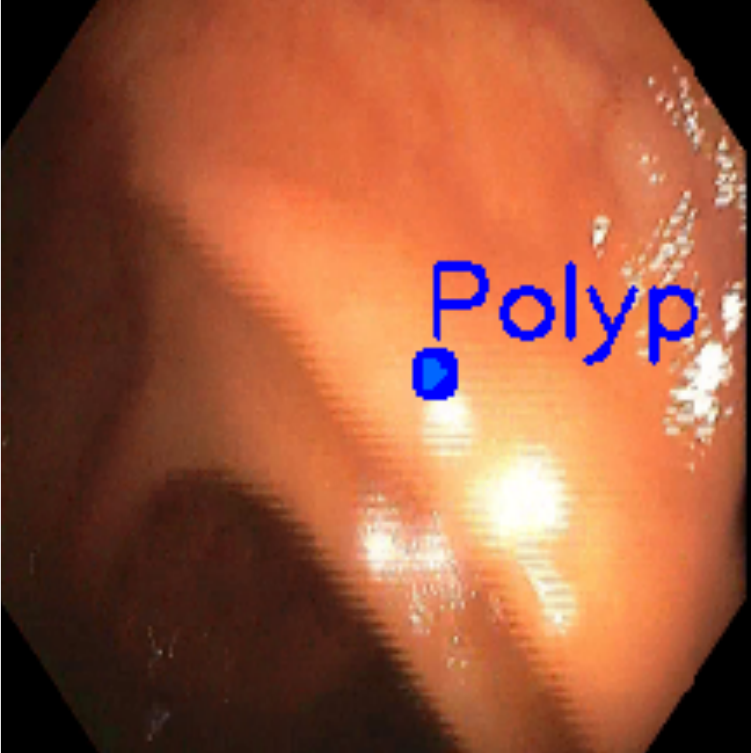}%
	\includegraphics[width=2cm,height=2cm]{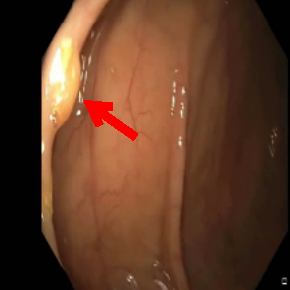}%
	\includegraphics[width=2cm,height=2cm]{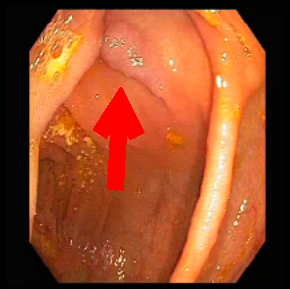}%
	\includegraphics[width=2cm,height=2cm]{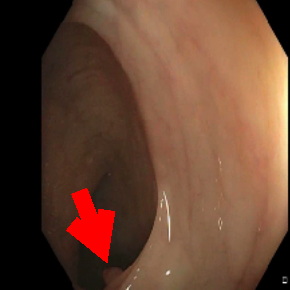}%
	\caption{Sample detection result on ASU-MAYO test dataset. The images are chosen to show variations of the polyp appearance and performance of our approach. The last row shows failure cases. The first three images show FP and the last three show FN. }
	\label{fig:comp}
\end{figure}
\subsection{Results}
We compare the proposed method with several methods from MICCAI 2015 challenge on polyp detection \cite{bernal2017comparative} and with the recent work \cite{yu2017integrating}. First, we show that the
performance of our model is superior comparing to these methods considering F1-score, F2-score, and recall performance metrics. Then we present various approaches to cast insight on polyp detection and limitation of our method. 
\newline
{\bf Comparison with the state-of-the-art:\quad} We compare out Y-Net method with the state-of-the-art methods listed in Table \ref{table:table1}. Polyp localize and spot team (PLS) on MICCAI 2015 challenge used a global hand-crafted image features followed by approximation of polyps with ellipses. CVC-CLINIC \cite{bernal2015wm} directly models polyps with shape descriptors. Both PLS and CVC-CLINIC\cite{bernal2015wm} use hand-crafted features followed by a classifier. ASU \cite{tajbakhsh2016automated} uses both hand-crafted features for estimating edge map followed by ensemble of CNN for final polyp localization. The rest of the methods are based on deep learning approaches, where OUS uses pre-trained AlexNet and CUMED proposed custom CNN for segmentation. The last method we compared with, Fusion \cite{yu2017integrating} uses 3D convolution with offline and online learning. Results in Table \ref{table:table1} demonstrate that our proposed method is superior to the other reference methods. The following observations can be made: (1) It is also interesting to observe that hand-crafted features have higher false positive rate than deep learning or hybrid approaches. We suspect this is because the polyps have varying shapes and appearance making it difficult for hand-crafting feature descriptor. (2) Our proposed method achieves the best performance in terms of F1-score, F2-score, and recall providing the maximum number of true detections  and the minimum number false detections. (3) Our result suggests that hybrid fine tuning a pre-trained network and training from scratch a mirrored network gives a better performance when the domain of data are different such as in natural vs. medical images. Sample visual results are shown in Fig. \ref{fig:comp}.
\newline
{\bf Detection latency:\quad}One of the challenges in clinical colonoscopy when performing fast inspection is that some of the polyps can appear in a few frames and could be missed in the subsequent video. Therefore, it is also important to
measure how quickly the first instance of the polyp can be detected by automated detection methods. Detection latency is the number of frames for our model to detect a polyp from the first appearance of a polyp in the scene, $\Delta T={t_2-t_1}$. The detection latency of our method for all test videos containing polyp is shown in Table \ref{table:table2}. It is important to mention that for all videos, we measure the latency from first appearance of polyp on ground truth data.

\begin{table}
		\caption{Comparison of different polyp detection methods on ASU-MAYO dataset. The result for PLS and OUS are published in \cite{bernal2017comparative}. It can be seen that Y-Net has the lowest number of FN, and highest number of TP. }
	\begin{center}
		\begin{tabular}{ c|c|c|c|c|c|c|cc }
			\hline \hline
			Method & TP &FP&FN&Prec[\%]&Rec[\%]&F1[\%]&F2[\%]\\
			\hline\hline
			PLS            & 1594 & 10103 &2719 &13.6 &36.9 &19.9 &27.5 \\
			CVC-CLINIC\cite{bernal2015wm}     & 1578 &  3456 &2735 &31.3 &36.6 &33.8 &35.4 \\
			OUS            & 2222 &   229 &2091 &90.6 &51.5 &65.7 &56.4 \\
			ASU\cite{tajbakhsh2016automated}			   & 2636 &   184 &1677 &\textbf{93.5} &61.1 &73.9 &65.7 \\
			CUMED 		   & 3081 &   769  &1232 &80.0 &71.4 &75.5 &73.0 \\
			Fusion\cite{yu2017integrating} 		   & 3062 &   414&1251 &88.1 &71.0 &78.6 &73.9 \\
			Y-Net(Ours)    & 3582 &   513&662 &87.4 &\textbf{84.4} &\textbf{85.9} &\textbf{85.0} \\
			\hline \hline
		\end{tabular}
	\end{center}

	\label{table:table1}
\end{table}

\subsection{Further insights and limitations}
We further compare our method with pre-trained VGG19 single encoder with skip connection. In this case, the architecture is similar to U-Net except the encoder network of U-Net is replaced with a pre-trained VGG19 model. We trained the model together with the decoder. Moreover, we trained U-Net from scratch in order to analyze the effect of having a pre-trained network as an encoder. The result is summarized in Table \ref{table:tableUnet}. As it can be seen, using a pre-trained network  as encoder part of U-Net improves the detection accuracy. Summing both pre-trained and un-trained encoder as proposed in this work results in the best F1 and F2-score for polyp detection. A unique characteristic of our method, compared with U-Net trained from scratch or with pre-trained encoder, is that it has a better compromise between recall and precision.

The last row of Fig. \ref{fig:comp}, shows some of false positives and false negatives of the proposed method. The sample frames are challenging in that there is a significant view point variation, shape, lighting and specular reflection. We have considered accounting for variation of light as well as contrast enhancement of the images as a preprocessing step. However, there is no significant difference in the accuracy.     

\begin{table}
	\caption{Detection latency for videos containing only polyps on ASU-MAYO test dataset.}
	\begin{center}
		\begin{tabular}{ c|c|c|c|c|c|c|c|c|cc }
			\hline \hline
			\multicolumn{9}{c} {Detection Latency}\\
			\hline\hline
			$\Delta T$            & 8 & 0 &0 &18 &0 &0 &0&0&0 \\
			Test video     & Vid5 &  Vid6 &Vid7 &Vid8 &Vid9 &Vid10 &Vid11&Vid12&Vid13 \\
			\hline \hline
		\end{tabular}
	\end{center}
	
	\label{table:table2}
\end{table}

\section{Conclusion}
We address polyp detection problem in colonoscopy videos by proposing a new deep encoder-decoder approach. Our method relies on two-encoder networks that use a pre-trained VGG19 weights in the first encoder and
random initialized weights in the latter. The two encoders are integrated with a novel sum-skip-concatenation operation. Moreover, we propose to use encoder based learning rate to efficiently use a pre-trained model. The approach is flexible in that, the pre-trained encoder used in this work can be replaced with other pre-trained models such as VGG16 or similar. The proposed fusion of pre-trained and untrained encoder network with sum-skip connection to decoder provides a new strategy to fill the gap between the large variation of testing data and the limited training data, which is a common challenge when employing supervised learning methods in medical image analysis tasks. Finally, experiment results on ASU-Mayo Clinic polyp database show that with the proposed multi-encoder framework, we achieved the best performance on F1 and F2 score metrics.  

\begin{table}
		\caption{Baseline comparison of pre-trained encoders: All the models are trained under the same data augmentation and common parameter settings}
	\begin{center}
		\begin{tabular}{ c|c|c|c|cc }
			\hline \hline
			Method & Prec[\%]&Rec[\%]&F1[\%]&F2[\%]\\
			\hline\hline
			U-Net(Trained from scratch)            &90.8 &39.2 &54.7 &44.2 \\
			U-Net(Pre-trained encoder, VGG19)              &\textbf{96.2} &68.2 &79.8 &72.4 \\
			Y-Net(Ours)                           &87.4 &\textbf{84.4} &\textbf{85.9} &\textbf{85.0} \\
			\hline \hline
		\end{tabular}
	\end{center}

	\label{table:tableUnet}
\end{table}
\bibliography{egbib}
\end{document}